# Surrogate-Assisted Reference Vector Adaptation to Various Pareto Front Shapes for Many-Objective Bayesian Optimization

Nobuo Namura
*Jiteki Lab*
Noda, Japan
nobuo.namura.gp@gmail.com

*Abstract*—We propose a surrogate-assisted reference vector adaptation (SRVA) method to solve expensive multi- and many-objective optimization problems with various Pareto front shapes. SRVA is coupled with a multi-objective Bayesian optimization (MBO) algorithm using reference vectors for scalarization of objective functions. The Kriging surrogate models for MBO is used to estimate the Pareto front shape and generate adaptive reference vectors uniformly distributed on the estimated Pareto front. We combine SRVA with expected improvement of penalty-based boundary intersection as an infill criterion for MBO. The proposed algorithm is compared with two other MBO algorithms by applying them to benchmark problems with various Pareto front shapes. Experimental results show that the proposed algorithm outperforms the other two in the problems whose objective functions are reasonably approximated by the Kriging models. SRVA improves diversity of non-dominated solutions for these problems with continuous, discontinuous, and degenerated Pareto fronts. Besides, the proposed algorithm obtains much better solutions from early stages of optimization especially in many-objective problems.

## I. INTRODUCTION

Real-world optimization problems are typically formulated as a multi-objective form with trade-offs among objective functions. In this study, we deal with the following multi-objective optimization problem:

$$\min_{\mathbf{x} \in S} \mathbf{f}(\mathbf{x}) = [f_1(\mathbf{x}) \quad \cdots \quad f_M(\mathbf{x})]^\mathrm{T}, \tag{1}$$

where $\mathbf{f}(\mathbf{x})$ is a vector of $M$ objective functions $f_i: S \to \mathbb{R}$ ($i = 1, \cdots, M$) with a $m$–dimensional design variable vector $\mathbf{x}$ in a design space $S$ which is a subset of $\mathbb{R}^m$. Pareto optimal solutions are obtained by solving (1) due to the trade-offs among the objective functions. Evolutionary algorithms (EAs) have been developed to solve the multi-objective optimization problems successfully and obtain diverse and converging non-dominated solutions (NDSs) [1,2]. EAs have been applied to many-objective optimization problems ($M \geq 4$) [3–7].

Another important feature of the real-world problems is computational time and cost for expensive function evaluation. Previous studies using computational fluid dynamics required couple of minutes [8] to one week [9] for the function evaluation of each solution. Thus, the number of function evaluation is limited by computational resources (e.g., 10–1,000 times). Surrogate models are frequently introduced for applying EAs to the expensive optimization problems. They are constructed for promptly estimating the values of the objective functions at any point in the design space from a set of sample points obtained by the expensive function evaluation. EAs search on the surrogate models using estimated values instead of expensive objective functions.

Bayesian optimization is one of the popular approaches to apply surrogate models to the expensive optimization problems [10]. The Kriging model (or Gaussian process model) is often used as the surrogate model in the Bayesian optimization. The Kriging model produces not only estimation of the objective function values but also uncertainty of the estimation. The Bayesian optimization utilizes the estimation and uncertainty to compute infill criteria (or acquisition function). Adding sample points (hereinafter, additional sample points) to the positions where the infill criteria have optimal values enables the Bayesian optimization to balance exploration and exploitation and efficiently solve the expensive optimization problems.

Some algorithms for multi-objective Bayesian optimization (MBO) adopt scalarizing functions, which transform a multi-objective problem into one or multiple single-objective problem(s), to compute infill criteria [11,12]. Many scalarizing functions use reference vectors which have the same dimension as the number of objective functions. Each component of the reference vector is a weight of the corresponding objective function. ParEGO [13] and MOEA/D-EGO [14] were typical algorithms for MBO using expected improvement (EI) as the infill criterion and Tchebycheff functions for scalarization. In these algorithms, the Kriging models were generated for each scalarized objective function. On the other hand, K-RVEA [15] generated the Kriging models for original objective functions to compute angle penalized distance as an infill criterion for each reference vector. EPBII [16] is another infill criterion utilizing the Kriging models of original objective functions to compute EI of penalty-based boundary intersection (PBI) in multi-dimensional objective space. In MBO with EPBII (MBO-EPBII), the Kriging models were used for not only computing the infill criterion but also estimating the Pareto front (PF). The PF was estimated through the approximation of each objective function to define nadir and utopia points for normalization of the objective space.

All of these methods adopted the simplex lattice-design (SLD) [17] and two-layered SLD [3] to generate the reference vectors uniformly distributing on a hyperplane in the objective space. However, the SLD method is unsuitable for the multi-objective problems with degenerated and discontinuous PFs and the many-objective problems. The SLD method suffers from a lack of NDSs for the degenerated and discontinuous PFs. Many of reference vectors do not intersect the PF and become waste of computational resources. Applying the SLD method to many-objective optimization problems results in enormous reference vectors to cover high-dimensional objective space. To tackle this issue, reference vector adaptation has been introduced into EAs [18-20]. RVEA [4] randomly regenerated reference vectors when no NDS was assigned to them. This approach worked well to improve diversity of NDSs for the problems with degenerated and discontinuous PFs. However, no reference vector adaptation method taking advantages of surrogate models has been proposed for MBO.

In this study, we propose a surrogate-assisted reference vector adaptation (SRVA) method for MBO to efficiently solve the multi- and many-objective optimization problems with various PF shapes including degenerated and discontinuous PFs. SRVA utilizes the PF shape estimated by the surrogate models of objective functions to determine distribution of reference vectors. We chose EPBII as the infill criterion for MBO coupled with SRVA (hereinafter MBO-EPBII-SRVA) because the PF estimation in MBO-EPBII can be easily diverted to SRVA. Besides, global optimization of hyperparameter with a genetic algorithm in MBO-EPBII can enhance accuracy of the PF estimation at the sacrifice of computational time. To validate effects of SRVA, MBO-EPBII-SRVA was compared with MBO-EPBII and K-RVEA by applying them to 4 types of benchmark problems, each of which had 3 and 6 objective functions. Comparison results showed that MBO-EPBII-SRVA obtained NDSs with better diversity than MBO-EPBII and K-RVEA with a limited number of function evaluation in most problems.

## II. KRIGING MODEL

For MBO and SRVA, we use the ordinary Kriging model which expresses the unknown function $f(\mathbf{x})$ as

$$f(\mathbf{x}) = \mu + \varepsilon(\mathbf{x}), \quad (2)$$

where $\mu$ is a global model and has a constant value and $\varepsilon(\mathbf{x})$ represents a local deviation from the global model that is defined as a Gaussian process following $N(0, \sigma^2)$. The correlation between $\varepsilon(\mathbf{x}^i)$ and $\varepsilon(\mathbf{x}^j)$ is strongly related to the distance between the corresponding points, $\mathbf{x}^i$ and $\mathbf{x}^j$ $(i, j = 1, \cdots, n)$. $n$ is the number of sample points (training data points). Various functions are available to define the correlation. In this study, we employ the Gaussian function with a weighted distance. The correlation between $\mathbf{x}^i$ and $\mathbf{x}^j$ is defined as

$$r(\mathbf{x}^i, \mathbf{x}^j) = \exp\left[-\sum_{k=1}^{m} \theta_k \left(x_k^i - x_k^j\right)^2\right], \quad (3)$$

where $\theta_k \geq 0$ is a hyperparameter and the $k$-th element of an $m$-dimensional weight vector $\boldsymbol{\theta}$. This hyperparameter provides the Kriging model with anisotropy and enhances its accuracy.

The Kriging predictor and uncertainty, which are proposed by Jones et al. [21], are expressed as

$$\hat{f}(\mathbf{x}) = \hat{\mu} + \mathbf{r}(\mathbf{x})^T \mathbf{R}^{-1}(\mathbf{y} - \mathbf{1}\hat{\mu}), \quad (4)$$

$$s^2(\mathbf{x}) = \hat{\sigma}^2 \left[1 - \mathbf{r}(\mathbf{x})^T \mathbf{R}^{-1} \mathbf{r}(\mathbf{x}) + \frac{\left(1 - \mathbf{1}^T \mathbf{R}^{-1} \mathbf{r}(\mathbf{x})\right)^2}{\mathbf{1}^T \mathbf{R}^{-1} \mathbf{1}}\right], \quad (5)$$

where $\hat{\mu}$ and $\hat{\sigma}^2$ are the estimated value of $\mu$ and $\sigma^2$, respectively. $\mathbf{R}$ denotes the $n \times n$ matrix whose $(i, j)$ entry is $r(\mathbf{x}^i, \mathbf{x}^j)$ while $\mathbf{r}(\mathbf{x})$ and $\mathbf{y}$ are $n$-dimensional vectors whose $i$-th elements are $r(\mathbf{x}, \mathbf{x}^i)$ and $f(\mathbf{x}^i)$, respectively. $\mathbf{1}$ is a $n$-dimensional vectors all of whose elements are 1. The unknown parameters in the Kriging model are $\boldsymbol{\theta}$, $\hat{\mu}$, and $\hat{\sigma}^2$. These are obtained by maximizing a likelihood function. $\hat{\mu}$ and $\hat{\sigma}^2$ are analytically derived as

$$\hat{\mu} = \frac{\mathbf{1}^T \mathbf{R}^{-1} \mathbf{y}}{\mathbf{1}^T \mathbf{R}^{-1} \mathbf{1}}, \quad (6)$$

$$\hat{\sigma}^2 = \frac{(\mathbf{y} - \mathbf{1}\hat{\mu})^T \mathbf{R}^{-1}(\mathbf{y} - \mathbf{1}\hat{\mu})}{n}. \quad (7)$$

The other parameter $\boldsymbol{\theta}$ is determined by maximizing the following log-likelihood function:

$$Ln(\boldsymbol{\theta}) = -\frac{1}{2}(n \ln \hat{\sigma}^2 + \ln|\mathbf{R}|). \quad (8)$$

In this study, a genetic algorithm is adopted to maximize (8).

## III. EXPECTED PBI IMPROVEMENT WITH SURROGATE-ASSISTED REFERENCE VECTOR ADAPTATION

MBO-EPBII-SRVA uses the Kriging models to estimate the PF shape for SRVA and compute expected PBI improvement (EPBII). A pseudo code of MBO-EPBII-SRVA is given in Algorithm 1. First, initial sample points are generated by the Latin hypercube sampling method [22] and their real objective functions are evaluated by expensive computation (Step 1). The main loop in Algorithm 1 consists of 8 steps. The Kriging model is constructed for each objective function with all sample points which have already been evaluated (Step 3). Then, the PF shape and nadir and utopia points are estimated for SRVA and normalization of the objective space by obtaining NDSs on the Kriging models (Steps 4 and 5). SRVA is conducted by using Algorithm 2 at Step 6 in Algorithm 1. Additional sample points are determined by the other two algorithms. Candidate solutions are found by maximizing EPBII for each reference vector with Algorithm 3 (step 7) while the additional sample points are selected from them by using Algorithm 4 (step 8). Finally, real objective functions of additional sample points are evaluated (Step 9), and sample points are updated (Step 10). The main loop is iterated until the number of sample points exceeds limitation. Details of Steps 4-8 in Algorithm 1 are described in the following subsections.

### A. Pareto Front Shape and Nadir/Utopia Points Estimation

The PF shape is estimated by obtaining NDSs on the Kriging models. We use NSGA-III [3] for this purpose to deal with many-objective problems while NSGA-II [1] was used in [16]. Function evaluation in NSGA-III is replaced by the estimation from the Kriging models. To generate well-distributed adaptive reference vectors, we need to obtain adequate number of NDSs by increasing the number of fixed reference vectors for NSGA-III. In this study, the number of fixed reference vectors for NSGA-III is at least five times more than that of adaptive reference vectors for EPBII. If practical EAs with reference vector adaptation are available, NSGA-III can be replace by them. Additionally, a single objective genetic algorithm is adopted to minimize each objective function and obtain an extreme solution. These solutions may be used to determine nadir and utopia points for objective function normalization.

Weak PFs should be eliminated from NDSs on the Kriging models to suitably normalize the objective space by using estimated PFs. This operation is especially important in artificial benchmark problems where multiple optimal solutions can be obtained in single objective minimization of each objective function (e.g., ZDT [23], DTLZ [24], WFG [25]). We adopt $\varepsilon$-dominance to eliminate the weak PF. In advance, objective functions of NDSs are tentatively normalized by minimum and maximum values in all NDSs, and all NDSs are added into a NDS set. One NDS is sequentially selected from the NDS set, and a small value $\varepsilon$ (0.01 in this study) is subtracted from each objective function of the other NDSs in the NDS set. If the selected NDS is dominated by the others subtracted by $\varepsilon$, this NDS is dropped from the NDS set. The NDS set after the iteration for all NDSs is used to determine the nadir and utopia points whose components consist of the maximum and minimum values of each objective function in the NDS set. Finally, the objective function values of the nadir and utopia points are added and subtracted by $\varepsilon$, respectively, and denormalized.

---

**Algorithm 1**: MBO-EPBII-SRVA

    **Input:** number of initial sample points $n_{init}$, number of maximum function evaluation $n_{max}$, number of additional sample points at each iteration $n_{add}$, number of reference vectors $N_{ref}$
    **Output:** NDSs among $n$ sample points $X$

1:    Generate initial sample points $X$ using the Latin hypercube sampling and evaluate objective functions $F$. Set number of sample points $n = n_{init}$
2:    while $n < n_{max}$
3:      Construct the Kriging models with $X$ and $F$
4:      Obtain NDSs on the Kriging models $\hat{X}$ and $\hat{F}$ using NSGA-III
5:      Delete weak Pareto optimal solution in $\hat{F}$ using $\varepsilon$-dominance and determine nadir and utopia points
6:      Generate $N_{ref}$ reference vectors $\Lambda$ with Algorithm 2
7:      Obtain $N_{ref}$ candidate solutions $X_c$ for $\Lambda$ through EPBII maximization using Algorithm 3
8:      Select $n_{add}$ additional sample points $X_a$ from $X_c$ using Algorithm 4
9:      Evaluate objective functions $F_a$ for $X_a$
10:    Update $X \leftarrow X \cup X_a$, $F \leftarrow F \cup F_a$, and $n = n + n_{add}$
11:   End while

---

*B. Surrogate-assisted Reference Vector Adaptation*

The adaptive reference vectors are generated by uniformly selecting NDSs of $N_{ref}$ (the number of reference vectors) on the Kriging models. Objective functions of sample points and NDSs on the Kriging models including the weak PF are normalized by the nadir and utopia points. Then, NDSs to be used as reference vectors, $\hat{\mathbf{f}}_{ref}^i$, are selected by computing maximin distances as follows:

$$\hat{\mathbf{f}}_{ref}^i = \underset{\hat{\mathbf{f}}^j \in \hat{F}}{\operatorname{argmax}}[d(\hat{\mathbf{f}}^j)], \quad (9)$$

$$d(\hat{\mathbf{f}}^j) = \min_{\mathbf{f} \in \Lambda \cup F}[\|\hat{\mathbf{f}}^j - \mathbf{f}\|], \quad (10)$$

where $F$ and $\hat{F}$ are archives for objective functions of sample points and NDSs on the Kriging models, respectively. $\Lambda$ is an archive for already selected NDSs to be used as reference vectors. $\mathbf{f}$ and $\hat{\mathbf{f}}^j$ are $M$-dimensional objective function vectors belonging to $\Lambda \cup F$ and $\hat{F}$, respectively. $\hat{\mathbf{f}}_{ref}^i$ is added into $\Lambda$, and a next NDS is selected by iteratively computing $\hat{\mathbf{f}}_{ref}^i$ until the number of NDSs in $\Lambda$ reaches $N_{ref}$. Note that we use the distance in the normalized objective space in (10) instead of that in the hyperplane for SLD. This enables the adaptive reference vectors to obtain uniformly distributed NDSs on PFs even if the PFs are degenerated and discontinuous. Using $F$ in (10) enhances to generate the adaptive reference vectors away from the already explored region in the objective space.

Selected NDSs in $\Lambda$ are divided into $n_{add}$ clusters using the k-means method [26] in the normalized objective space as well as the NDS selection. $n_{add}$ is the number of additional sample points at each iteration. Cluster centroids of the k-means method are determined only with NDSs inside a hypercube defined by the nadir and utopia points. The other NDSs on the weak PF are assigned to the nearest centroids. One additional sample point is selected from each cluster at Step 8 in Algorithm 1. Classified NDSs in $\Lambda$ are transformed into unit vectors, and we use them as adaptive reference vectors. The process for SRVA described above is summarized in Algorithm 2.

---

**Algorithm 2**: SRVA

    **Input:** number of additional sample points at each iteration $n_{add}$, number of reference vectors $N_{ref}$, NDSs on the Kriging models $\hat{F}$, objective functions of sample points $F$, nadir and utopia points
    **Output:** classified reference vectors in $\Lambda$

1:    Normalize $F$ and $\hat{F}$ using nadir and utopia points
2:    Initialize an empty archive for reference vectors $\Lambda \leftarrow \emptyset$
3:    For $i = 1, \cdots, N_{ref}$
4:      Select one NDS $\hat{\mathbf{f}}_{ref}^i$ from $\hat{F}$ using (9) and (10)
5:      Update $\Lambda \leftarrow \Lambda \cup \{\hat{\mathbf{f}}_{ref}^i\}$
6:    End for
7:    Divide objective function vectors in $\Lambda$ into $n_{add}$ clusters using the k-means method
8:    Generate reference vectors by transforming the objective function vectors in $\Lambda$ into unit vectors

## C. Expected PBI Improvement

EPBII is the expected value of the PBI improvement (PBII) and derived from the predictor and uncertainty from the Kriging models $\mathcal{N}(\hat{f}_k(\mathbf{x}), s_k^2(\mathbf{x}))$ $(k = 1, \cdots, M)$. Along the $i$-th reference vector $\boldsymbol{\lambda}^i$ ($\|\boldsymbol{\lambda}^i\| = 1$, $i = 1, \cdots, N_{ref}$), a PBI function $g(\mathbf{f}, \boldsymbol{\lambda}^i)$ for a certain point $\mathbf{f}$ in the normalized objective space is defined as follows:

$$g(\mathbf{f}, \boldsymbol{\lambda}^i) = d_1 + \theta_{PBI} d_2, \quad (11)$$
$$d_1 = \|\mathbf{f}^T \boldsymbol{\lambda}^i\|, \quad (12)$$
$$d_2 = \|\mathbf{f} - d_1 \boldsymbol{\lambda}^i\|, \quad (13)$$

where $\theta_{PBI}$ is a penalty parameter, and we use $\theta_{PBI} = 1$ as well as [16].

EPBII is computed by numerical integration of PBII and the $M$-dimensional probability density functions $\varphi(f_k)$ denoted by $\mathcal{N}(\hat{f}_k(\mathbf{x}), s_k^2(\mathbf{x}))$ as follows:

$$EPBII(\mathbf{x}, \boldsymbol{\lambda}^i, F) \quad (14)$$
$$= \begin{cases} \int_{-\infty}^{\infty} \cdots \int_{-\infty}^{\infty} PBII(\mathbf{f}, \boldsymbol{\lambda}^i, F)\varphi(f_1)\cdots\varphi(f_M)df_1\cdots df_M \\ \qquad\qquad\qquad\qquad (T(\hat{\mathbf{f}}(\mathbf{x}), \boldsymbol{\lambda}^i) \geq 0) \\ T(\hat{\mathbf{f}}(\mathbf{x}), \boldsymbol{\lambda}^i) \qquad (T(\hat{\mathbf{f}}(\mathbf{x}), \boldsymbol{\lambda}^i) < 0) \end{cases}$$

$$PBII(\mathbf{f}, \boldsymbol{\lambda}^i, F) = \max[g_{ref}^i - g(\mathbf{f}, \boldsymbol{\lambda}^i), 0], \quad (15)$$
$$g_{ref}^i = \min_{\mathbf{f} \in F_T^i}[g(\mathbf{f}, \boldsymbol{\lambda}^i)], \quad (16)$$
$$F_T^i = \{\mathbf{f} \in F | T(\mathbf{f}, \boldsymbol{\lambda}^i) \geq 0\},$$
$$T(\mathbf{f}, \boldsymbol{\lambda}^i) = d_1 - \theta_{ref} d_2. \quad (17)$$

where $\hat{\mathbf{f}}(\mathbf{x})$ is a $M$-dimensional vector whose $k$-th element is $\hat{f}_k(\mathbf{x})$. $T(\mathbf{f}, \boldsymbol{\lambda}^i)$ is a territory function which defines search space (territory) in the objective space assigned to each reference vector. As shown in (14), EPBII for $\boldsymbol{\lambda}^i$ is computed as the expected value of PBII only inside the territory ($T(\mathbf{f}, \boldsymbol{\lambda}^i) \geq 0$) while the territory function leads solutions outside the territory ($T(\mathbf{f}, \boldsymbol{\lambda}^i) < 0$) to the inside. PBII for $\boldsymbol{\lambda}^i$ is computed as the difference between reference PBI $g_{ref}^i$, which is the minimum PBI for the sample points inside the territory $F_T^i$, and PBI at a certain point $\mathbf{f} = [f_1, \cdots, f_M]^T$. One hundred points of $\mathbf{f}$ are generated by a Monte Carlo sampling following $\mathcal{N}(\hat{f}_k(\mathbf{x}), s_k^2(\mathbf{x}))$ for the numerical integration in (14). $\theta_{ref}$ is a parameter to adjust the size of the territory defined as

$$\theta_{ref} = \frac{\sqrt{2}}{d_{min}}, \quad (18)$$

$$d_{min} = \frac{1}{N_{ref}} \sum_{i=1}^{N_{ref}} \min_{j \neq i}[d^{ij}], \quad (19)$$

$$d^{ij} = \|\tilde{\boldsymbol{\lambda}}^i - \tilde{\boldsymbol{\lambda}}^j\|. \quad (20)$$

$\tilde{\boldsymbol{\lambda}}^i$ and $\tilde{\boldsymbol{\lambda}}^j$ are reference vectors projected onto the hyperplane where a sum of vector elements is one. $\theta_{ref}$ in (18) is modified from [16] to deal with the adaptive reference vectors.

---

**Algorithm 3**: EPBII Maximization

**Input:** classified reference vectors in $\Lambda$, number of reference vectors $N_{ref}$, NDSs on the Kriging models $\hat{X}$ and $\hat{F}$, objective functions of sample points $F$, nadir and utopia points, Kriging models

**Output:** candidate solutions $X_c$ with their EPBII, distances $d^{ij}$ and the mean minimum distance $d_{min}$ among reference vectors in $\Lambda$ projected onto the hyperplane, $F$ inside each territory $F_T^i$

1: Compute $\theta_{ref}$, $d_{min}$, and $d^{ij}$ from $\Lambda$ using (18)-(20)
2: Initialize an empty archive for candidate solutions $X_c \leftarrow \emptyset$
3: For $i = 1, \cdots, N_{ref}$
4:    Compute reference PBI $g_{ref}^i$ for the $i$-th reference vector and obtain $F_T^i$
5:    Add one of NDSs in $\hat{X}$, whose $\hat{F}$ is closest to the $i$-th reference vector, into $X_c$ as initial population
6: End for
7: Run MOEA/D to maximize EPBII for $\Lambda$ and obtain optimized candidate solutions $X_c$

---

EPBII for each reference vector is maximized by Algorithm 3 to obtain optimized candidate solutions. $\theta_{ref}$ and the reference PBI for the current MBO iteration are computed beforehand because these values are constant through EPBII maximization. Additionally, we use NDSs on the Kriging models obtained at Step 4 in Algorithm 1 as initial population for the EPBII maximization. $N_{ref}$ NDSs closest to each reference vector are selected as the initial population. In this study, MOEA/D [2] is adopted to simultaneously maximize EPBII for each reference vector while EPBII was individually maximized as $N_{ref}$ single objective problems in [16].

## D. Selection of Additional Sample Points

Additional sample points are selected from the candidate solutions according to their fitness derived from EPBII, niche counts, and Pareto ranking. The niche count $nc^i$ of the $i$-th candidate solution maximizing EPBII for $\boldsymbol{\lambda}^i$ is computed as

$$nc^i = \sum_{j=1}^{N_{ref}} \frac{n_{NDS}^j}{h(d^{ij}/d_{min}) + 1}, \quad (21)$$

$$h(x) = \begin{cases} 2x - 1 & (x > 1) \\ x^2 & (x \leq 1) \end{cases}, \quad (22)$$

where $h(x)$ is a correction function and $n_{NDS}^j$ is the number of NDSs belonging to $F_T^j$. If multiple sample points are added in the same MBO iteration, $n_{NDS}^j$ includes the number of additional sample points which have already been added in the current MBO iteration and are inside the territory of $\boldsymbol{\lambda}^j$. A reference vector passing near NDSs has a high niche count. Thus, selecting the candidate solutions with low niche counts improves diversity of NDSs among the sample points.

Using $nc^i$ and the Pareto ranking of $i$-th candidate solution $rank^i$, EPBII is converted into the following fitness:

$$fitness^i = \frac{EPBII(\mathbf{x}_c^i, \boldsymbol{\lambda}^i, F)}{nc^i \cdot rank^i}, \quad (23)$$

where $\mathbf{x}_c^i$ is a candidate solution maximizing EPBII for $\boldsymbol{\lambda}^i$. One candidate solution with the highest fitness in each cluster is selected as additional sample points. This selection is conducted for each cluster one by one while updating $n_{NDS}^j$. The detailed process is summarized in Algorithm 4.

## IV. NUMERICAL EXPERIMENT

In this section, we apply MBO-EPBII-SRVA to 4 types of benchmark problems to investigate effects of SRVA and its performance in multi- and many-objective problems with various PF shapes. MBO-EPBII-SRVA was compared with MBO-EPBII and K-RVEA in these problems. MBO-EPBII was implemented by replacing Step 6 in Algorithm 1 by the two-layered SLD and not the same as implementation in [16]. MBO-EPBII(-SRVA) was implemented in Python while we used K-RVEA implemented in MATLAB and provided through GitHub repository of the author of [15]. The Python implementation of MBO-EPBII-SRVA is available on our GitHub repository[1].

### A. Experimental Setup

DTLZ1, 2, 5, 7 benchmark problems with 3 and 6 objective functions ($M = 3$ and 6) were used in numerical experiments. The number of design variables was 10 in all problems. DTLZ1 and 2 were selected to examine effects of approximation accuracy of the Kriging model on SRVA. DTLZ1 is a typical problem difficult to approximate while DTLZ2 is easy to approximate. DTLZ5 and 7 were used to examine the performance of SRVA on degenerated and discontinuous PFs, respectively.

---

**Algorithm 4**: Selection of Additional Sample Points

**Input:** number of additional sample points at each iteration $n_{add}$, candidate solutions $X_c$ with their EPBII, classified reference vectors in $\Lambda$, objective functions of sample points $F$, $F$ inside each territory $F_T^j$, distances $d^{ij}$ and the mean minimum distance $d_{min}$ among reference vectors in $\Lambda$, Kriging models

**Output:** additional sample points $X_a$

1: Evaluate objective functions $\hat{F}_c$ for $X_c$ using the Kriging models and perform Pareto ranking for $\hat{F}_c$
2: Count up the number of NDSs belonging to each $F_T^j$ and obtain $n_{NDS}^j$
3: Initialize an empty archive for additional sample points $X_a \leftarrow \emptyset$
4: For $k = 1, \cdots, n_{add}$
5:    Compute niche count and fitness of candidate solutions in $X_c$ belonging to the $k$-th cluster using (21)-(23)
6:    Add the candidate solution with the highest fitness in the $k$-th cluster $\mathbf{x}_a^k$ into $X_a$
7:    Add estimated objective functions $\hat{F}_c$ at $\mathbf{x}_a^k$ into corresponding $F_T^j$ and update $n_{NDS}^j$
8: End for

---

[1] https://github.com/Nobuo-Namura/MBO-EPBII-SRVA

Parameters of the three algorithms used in the experiments are summarized as follows:

1) Number of initial sample points: $n_{init} = 30$
2) Number of maximum function evaluation: $n_{max} = 300$
3) Number of additional sample points at each iteration: $n_{add} = 10$
4) Number of independent runs = 11
5) Number of reference vectors: $N_{ref} = 91$ and 112 for 3 and 6 objective problems, respectively
6) Design factors $(H_1, H_2)$ of the two-layered SLD used in MBO-EPBII and K-RVEA: $(H_1, H_2) = (12, 0)$ and $(3, 3)$ to meet $N_{ref} = 91$ and 112, respectively.
7) Design factors $(H_1, H_2)$ of the two-layered SLD for NSGA-III in MBO-EPBII(-SRVA): $(H_1, H_2) = (30, 0)$ and $(6, 5)$ resulting in 496 and 716 reference vectors for 3 and 6 objective problems, respectively
8) Number of generations in NSGA-III = 200
9) Number of generations in MOEA/D for MBO-EPBII(-SRVA) = 50

The other parameters used in K-RVEA were set to its default.

In order to compare the performance of three algorithms, we evaluated hypervolume and inverted generational distance plus (IGD+) [27] for sample points at every iteration. Reference points for the hypervolume and the number of reference points for IGD+ are shown in Table I and II, respectively.

TABLE I. REFERENCE POINTS FOR HYPERVOLUME COMPUTATION

| Problem | 3 objectives | 6 objectives |
|---|---|---|
| DTLZ1 | [150, 150, 150] | [50, 50, 50, 50, 50, 50] |
| DTLZ2 | [1.1, 1.1, 1.1] | [1.1, 1.1, 1.1, 1.1, 1.1, 1.1] |
| DTLZ5 | [1.1, 1.1, 1.1] | [1.1, 1.1, 1.1, 1.1, 1.1, 1.1] |
| DTLZ7 | [1.1, 1.1, 6.1] | [1.1, 1.1, 1.1, 1.1, 1.1, 12.1] |

TABLE II. NUMBER OF REFERENCE POINTS FOR IGD+ COMPUTATION

| Problem | 3 objectives | 6 objectives |
|---|---|---|
| DTLZ1 | 1326 | 8568 |
| DTLZ2 | 1326 | 8568 |
| DTLZ5 | 2000 | 8000 |
| DTLZ7 | 2401 | 7776 |

### B. Results and Discussion

Statistics of the hypervolume and IGD+ at the end of MBO in 11 independent runs are summarized in Tables III and IV, respectively. Results of the Wilcoxon signed rank test at a significance level of 0.01 are shown as three types of symbols $+, -,$ and $\approx$. The symbols $+$ and $-$ indicate MBO-EPBII-SRVA has significantly larger and smaller values than the other algorithm (MBO-EPBII or K-RVEA) while the symbol $\approx$ indicates there is no significant difference. Correction for multiple comparisons was not applied. The best values for each problem are highlighted.

Tables III and IV show that MBO-EPBII-SRVA achieved the best performance among three algorithms except for DTLZ1 which has many local optimums and is hard to approximate.

Comparison between MBO-EPBII-SRVA and MBO-EPBII revealed that SRVA can improve the performance of MBO for the problems which can be approximated by the Kriging models (DTLZ2, 5, 7). Besides, SRVA did not disturb the MBO performance even for the problems difficult to approximate since there is no significant difference between MBO-EPBII-SRVA and MBO-EPBII in DTLZ1. K-RVEA outperformed the other two algorithms in DTLZ1 because sample points can be added to the position maximizing the Kriging uncertainty regardless of the approximation accuracy of the Kriging models.

For the problems difficult to approximate, surrogate-assisted algorithms which do not use objective function approximation are available [28,29]. Compared with these algorithms, MBO-EPBII-SRVA is useful for the problems which can be approximated by the Kriging models.

Mean hypervolume for each MBO iteration is shown in Fig. 1 where shading represents 25 and 75 percentiles. MBO-EPBII-SRVA showed much better performance than K-RVEA from early stages of optimization. This feature was enhanced in 6 objective problems. The hypervolume of MBO-EPBII-SRVA increased from that of MBO-EPBII even in DTLZ2 with continuous PFs where well-distributed NDSs can be easily obtained by fixed reference vectors of SLD. In MBO-EPBII, many sample points were added to the position where one of the objective functions was zero due to the weak PF on the Kriging models. SRVA led additional sample points to the true PF, and diversity of NDSs was improved in MBO-EPBII-SRVA as shown in Fig. 2 where NDSs of 3 objective DTLZ2 obtained in the run with median hypervolume are visualized. The hypervolume in 6 objective DTLZ2 was greatly increased by introducing SRVA because SRVA generated reference vectors away from the already explored region in the objective space.

TABLE III. STATISTICS OF HYPERVOLUME OBTAINED BY THREE ALGORITHMS

| Problem | $M$ | MBO-EPBII-SRVA | | | | | MBO-EPBII | | | | | K-RVEA | | | | |
|---|---|---|---|---|---|---|---|---|---|---|---|---|---|---|---|---|
| | | mean | std† | min | max | | mean | std† | min | max | | mean | std† | min | max | |
| DTLZ1 | 3 | 2.153 | 0.331 | 1.449 | 2.574 | ≈ | 2.074 | **0.297** | **1.618** | 2.575 | ≈ | **2.467** | 0.522 | 1.399 | **2.960** | $\times 10^6$ |
| | 6 | 0.653 | 0.188 | 0.348 | 0.945 | ≈ | 0.701 | 0.339 | 0.000 | 1.050 | − | **1.186** | **0.151** | **0.966** | **1.390** | $\times 10^{10}$ |
| DTLZ2 | 3 | **0.712** | **0.007** | **0.699** | **0.723** | + | 0.686 | 0.012 | 0.664 | 0.707 | + | 0.702 | 0.008 | 0.682 | 0.710 | |
| | 6 | **1.361** | **0.016** | **1.334** | **1.382** | + | 1.157 | 0.039 | 1.086 | 1.195 | + | 1.261 | 0.083 | 1.106 | 1.368 | |
| DTLZ5 | 3 | **0.426** | **0.003** | **0.418** | **0.429** | + | 0.396 | 0.007 | 0.381 | 0.402 | + | 0.388 | 0.008 | 0.373 | 0.400 | |
| | 6 | 0.390 | 0.025 | 0.339 | **0.422** | ≈ | 0.382 | 0.015 | 0.356 | 0.410 | ≈ | **0.402** | 0.013 | **0.368** | 0.415 | |
| DTLZ7 | 3 | **2.011** | **0.007** | **2.001** | **2.024** | + | 1.967 | 0.018 | 1.925 | 1.993 | + | 1.917 | 0.018 | 1.897 | 1.951 | |
| | 6 | **4.377** | **0.036** | **4.329** | **4.448** | + | 3.974 | 0.055 | 3.906 | 4.067 | + | 3.138 | 0.412 | 2.424 | 3.660 | |

†Standard deviation.

TABLE IV. STATISTICS OF IGD+ OBTAINED BY THREE ALGORITHMS

| Problem | $M$ | MBO-EPBII-SRVA | | | | | MBO-EPBII | | | | | K-RVEA | | | | |
|---|---|---|---|---|---|---|---|---|---|---|---|---|---|---|---|---|
| | | mean | std† | min | max | | mean | std† | min | max | | mean | std† | min | max | |
| DTLZ1 | 3 | 75.96 | 17.19 | 57.88 | 112.5 | ≈ | 74.52 | **16.95** | 50.60 | **101.1** | ≈ | **61.26** | 21.19 | **39.23** | 111.1 | |
| | 6 | 30.66 | 7.15 | 21.90 | 46.05 | ≈ | 32.21 | 12.23 | 20.03 | 55.83 | + | **19.47** | **6.83** | **11.79** | **32.29** | |
| DTLZ2 | 3 | **0.036** | 0.003 | **0.031** | **0.041** | − | 0.044 | 0.004 | 0.036 | 0.050 | − | 0.040 | **0.003** | 0.036 | 0.046 | |
| | 6 | **0.166** | **0.006** | 0.153 | **0.174** | − | 0.202 | 0.009 | 0.187 | 0.216 | ≈ | 0.174 | 0.025 | **0.145** | 0.230 | |
| DTLZ5 | 3 | **0.010** | **0.002** | **0.008** | **0.015** | − | 0.032 | 0.006 | 0.025 | 0.042 | − | 0.036 | 0.005 | 0.027 | 0.043 | |
| | 6 | **0.047** | 0.010 | **0.034** | **0.061** | − | 0.073 | 0.019 | 0.045 | 0.111 | ≈ | 0.052 | **0.008** | 0.043 | 0.067 | |
| DTLZ7 | 3 | **0.018** | **0.001** | **0.017** | **0.020** | − | 0.030 | 0.005 | 0.025 | 0.044 | − | 0.045 | 0.005 | 0.033 | 0.053 | |
| | 6 | **0.297** | **0.041** | **0.261** | **0.412** | − | 0.463 | 0.093 | 0.311 | 0.617 | − | 0.388 | 0.067 | 0.320 | 0.560 | |

†Standard deviation.

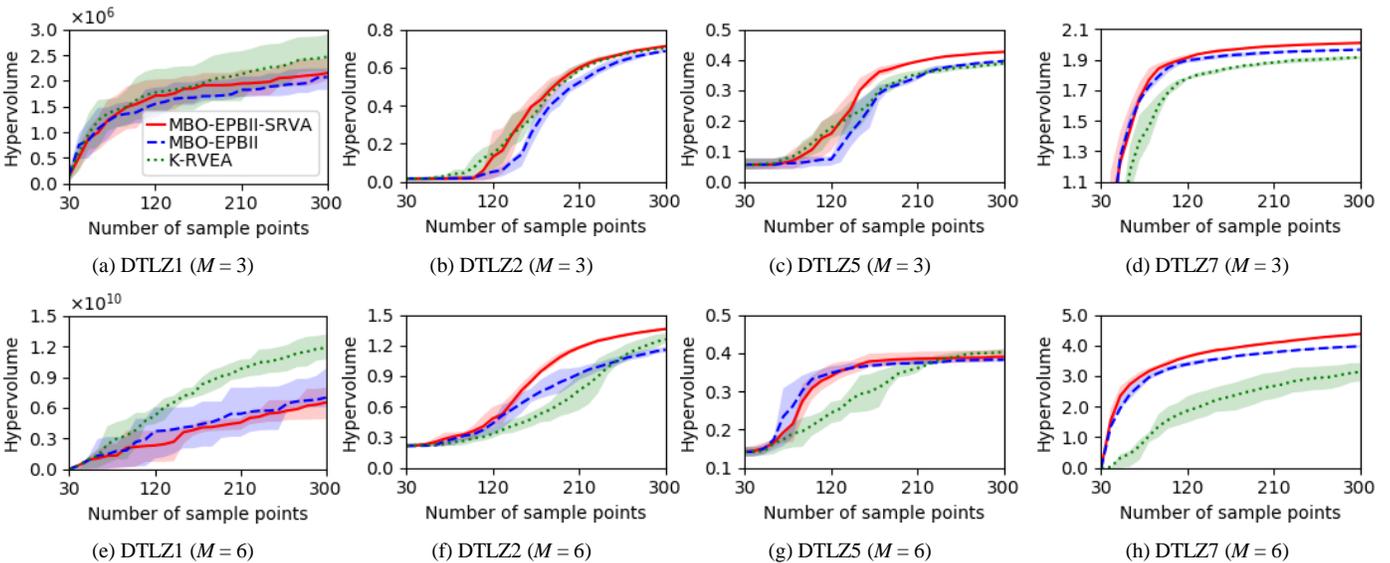

Fig. 1. Mean hypervolume obtained by three algorithms for each MBO iteration. Shading shows 25 and 75 percentiles.

MBO-EPBII-SRVA obtained many NDSs close to the degenerated PF of 3 objective DTLZ5 as shown in Fig. 3. This result indicates that SRVA helps MBO efficiently solve the problems with degenerated PFs. However, the performance of three algorithms was comparable in 6 objective DTLZ5. This may be caused by NSGA-III used in Algorithm 1. SRVA cannot generate well-distributed reference vectors if the number of NDSs on the estimated PF is small. The degenerated PF of DTLZ5 was hard to explore with fixed reference vectors generated by two-layered SLD in NSGA-III when the number of objective function is large. We can improve the performance of MBO-EPBII-SRVA on many-objective DTLZ5 by replacing NSGA-III by another EA with adaptive reference vectors.

SRVA most effectively worked in DTLZ7 with the discontinuous PF. As shown in Fig. 4, MBO-EPBII-SRVA obtained the adequate number of NDSs on all part of the discontinuous PF in 3 objective DTLZ7 while NDSs obtained by MBO-EPBII and K-RVEA were non-uniformly distributed on each part of the PF. As a result, MBO-EPBII-SRVA with 120 sample points achieved the comparable hypervolume obtained by K-RVEA with 300 sample points. In 6 objective DTLZ7, the difference between MBO-EPBII-SRVA and K-RVEA was emphasized due to the small number of fixed reference vectors intersecting the PF in K-RVEA. SRVA efficiently assigned limited number of reference vectors to the entire high-dimensional objective space.

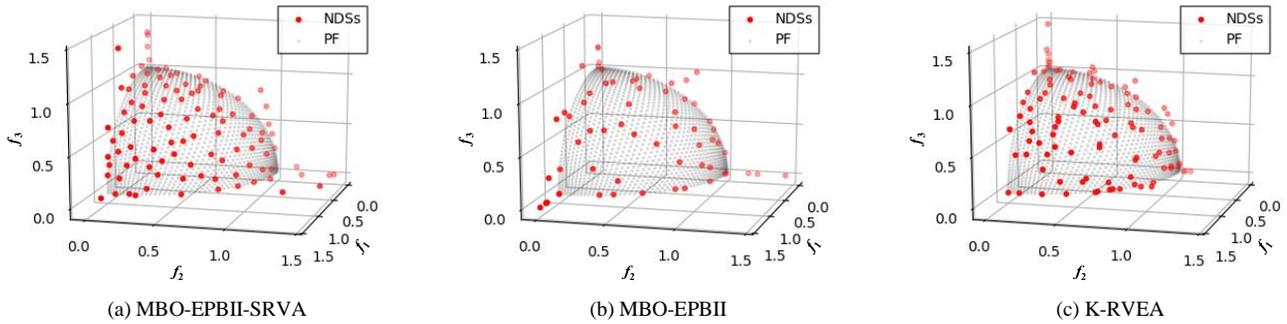

Fig. 2. NDSs of 3 objective DTLZ2 obtained in the run with median hypervolume in each algorithm.

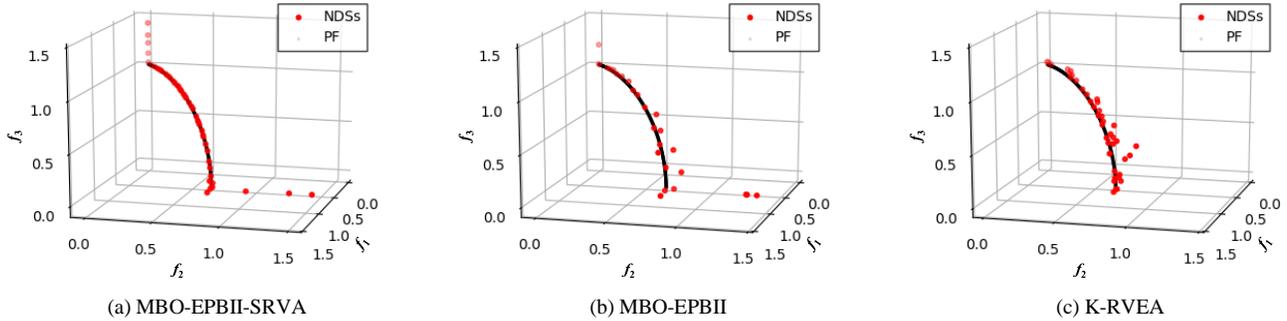

Fig. 3. NDSs of 3 objective DTLZ5 obtained in the run with median hypervolume in each algorithm.

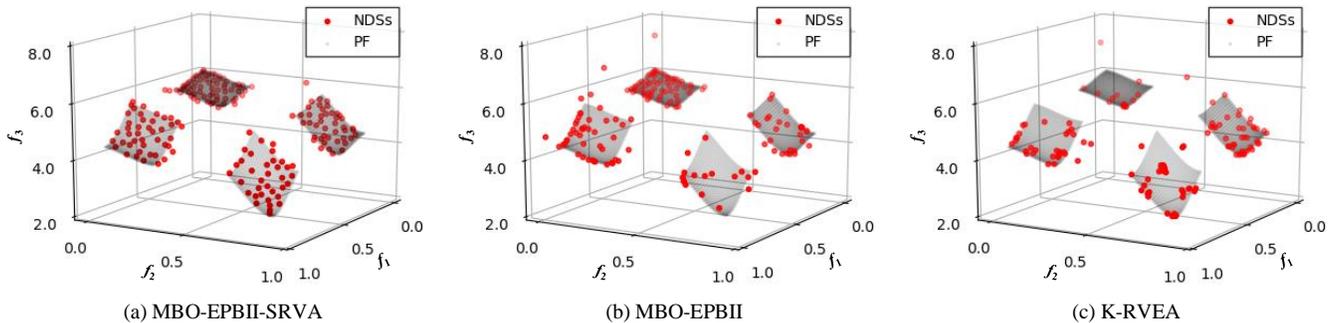

Fig. 4. NDSs of 3 objective DTLZ7 obtained in the run with median hypervolume in each algorithm.

## V. Conclusions and Future Work

In this study, a surrogate-assisted reference vector adaptation (SRVA) was proposed to solve the multi- and many-objective optimization problems with various Pareto front shapes. We combined SRVA with multi-objective Bayesian optimization with expected PBI improvement (MBO-EPBII) and named it MBO-EPBII-SRVA. MBO-EPBII-SRVA was compared with two MBO algorithms (MBO-EPBII and K-RVEA) in four types of benchmark problems, each of which had 3 and 6 objective functions. Comparison between MBO-EPBII-SRVA and MBO-EPBII showed that SRVA improved diversity of non-dominated solutions for the problems with continuous, discontinuous, and degenerated Pareto fronts if the objective functions were reasonably approximated by the surrogate models. MBO-EPBII-SRVA showed comparable performance to MBO-EPBII even in the problems difficult to approximate. Besides, MBO-EPBII-SRVA obtained much better solutions than K-RVEA from early stages of optimization especially in many-objective problems. As future work, we will replace NSGA-III used in MBO-EPBII-SRVA by another evolutionary algorithm with adaptive reference vectors to improve the performance of SRVA.